# Google street view and deep learning: a new ground truthing approach for crop mapping

Yulin Yan[a] and Youngryel Ryu[a,b*]

[a] *Interdisciplinary Program in Agricultural and Forest Meteorology, Seoul National University, South Korea.*

[b] *Department of Landscape Architecture and Rural Systems Engineering, Seoul National University, South Korea.*

∗Corresponding author at: Department of Landscape Architecture and Rural Systems Engineering, Seoul National University, Seoul 151-921, South Korea. E-mail address: yryu@snu.ac.kr (Y. Ryu).

**Abstract**

Ground referencing is essential for supervised crop mapping. However, conventional ground truthing involves extensive field surveys and post processing, which is costly in terms of time and labor. In this study, we applied a convolutional neural network (CNN) model to explore the efficacy of automatic ground truthing via Google street view (GSV) images in two distinct farming regions: central Illinois and southern California. We demonstrated the feasibility and reliability of the new ground referencing technique further by performing pixel-based crop mapping with vegetation indices as the model input. The results were evaluated using the United States Department of Agriculture (USDA) crop data layer (CDL) products. From 8,514 GSV images, the CNN model screened out 2,645 target crop images. These images were well classified into crop types, including alfalfa, almond, corn, cotton, grape, soybean, and pistachio. The overall GSV image classification accuracy reached 93% in California and 97% in Illinois. We then shifted the image geographic coordinates using fixed empirical coefficients to produce 8,173 crop reference points including 1,764 in Illinois and 6,409 in California. Evaluation of these new reference points with CDL products showed satisfactory coherence, with 94–97% agreement. CNN-based mapping also captured the general pattern of crop type distributions. The overall differences between CDL products and our mapping results were 4% in California and 5% in Illinois. Thus, using these deep learning and GSV image techniques, we have provided an efficient and cost-effective alternative method for ground referencing and crop mapping.

**Keywords**: Google Street View, ground referencing, convolutional neural network, crop mapping

# 1. Introduction

Supervised crop type classification requires extensive ground truthing for training and validation (Foody and Mathur, 2004). The quality and quantity of reference data used to label crop types fundamentally affects classification accuracy (Foody et al., 2016; Kavzoglu, 2009). The sources for these reference data commonly include filed surveys, census data, or visual interpretation of remote sensing products (Dong et al., 2016; Jia et al., 2013; Wardlow et al., 2007). However, conventional ground truthing is time consuming, labor-intensive and costly (Hanuschak and Mueller, 2002). Lacking a low-cost and efficient reference producing method pervasively result in limited ground reference and hinder crop classification but still attracted few attentions.

Annual cropland data layer (CDL) products released by the United States Departments of Agriculture (USDA) are the most successful ground truthing products to date in terms of nationwide coverage, annual updating frequency, and high mapping accuracy (85–95%) (Boryan et al., 2011). Since 2008, many studies have relied on CDL products for geo-referenced and crop-specific maps at a spatial resolution of 30–56 m (Howard and Wylie, 2014; King et al., 2017; Skakun et al., 2017; Torbick et al., 2018). However, many regions and countries are incapable to produce long-term, large-scale, and accurate in situ observations. Besides, consistency issues of the in situ data between countries usually limit its application (King et al., 2017). The extremely limited amount of publicly available ground reference data is the critical obstacle of crop type mapping, especially for developing countries. Therefore, frequent updating of large-scale crop maps with high spatial resolution remains a great challenge (You et al., 2014).

The efficient acquisition and sharing of sufficiently high-quality ground truthing observations are therefore goals for both scientific and practical application. Except the CDL products, a recent crowdsource-based project attempted to initiate collaborations between scientists and citizens to collect and distribute geo- and time-referenced filed photographs on a global scale (Xiao et al., 2011). As of September 2019, the resulting data portal stores more than 180,000 land cover images. However, there is no quality control of these archived images; users can upload incorrect images, and spatial and temporal data gap are also considerable. Another crowdsourced project, GeoWiki recruited volunteers to review and improve global land cover map products using the Google Earth platform (Fritz et al., 2009). Citizen science thus provides unconventional solutions for generating ground truth data. However, encouraging citizens to participate and quality assurance remain challenges.

Google Street View (GSV) was established in 2007; it provides vast amounts of panoramic images at a global scale, with detailed geographic coordinates and time information. These data hold great potential for scientific studies. For example, Ringland et al. (2019) applied a pre-trained convolutional neural network (CNN) model to GSV images to characterize food production along roads. Gebru et al. (2017) also employed a CNN model to analyze socioeconomic profiles across 200 American cities using GSV images. GSV images are an unprecedented high-quality data source that directly captures land cover information; however, it has been under-exploited. Deep learning, especially in CNN models, has been successfully applied for image classification tasks (Ciresan et al., 2011; Razavian et al., 2014). Applying a CNN model to GSV images could be a promising alternative for the efficient production of large amounts of ground reference data. GSV images are available only along roadsides; therefore, it remains unclear whether they are feasible and robust as proxy for generating broadly representative ground reference data for crop type mapping.

The objective of this study was to develop an off-the-shelf crop type referencing and classification method, which can be adapted for any region worldwide and can be upscaled efficiently to large areas where GSV images are available. We trained a CNN model and applied it to GSV images to produce a crop type reference dataset. We then modified the structure of the CNN model and applied it jointly with Landsat surface reflectance products to produce a regional-scale crop type map.

## 2. Methods and Materials

2.1 Study area and general design

Two areas were included in this study. Study area I (119° 26´ – 119° 17´ W, 35° 26´ – 35° 34´ N) is located in southern California, USA (Fig. 1a). A rectangular area (total area, 18,088 ha) is same with 13.6 km × 13.3 km. According to the CDL products, this area contains diverse crop types; including alfalfa, almonds, corn, cotton, grape, pistachios, and wheat, as well as few other crops varying annually. Winter wheat and summer maize are commonly cultivated in this region. Study area II (89° 47´ – 89° 24´ W, 40° 08´ – 40° 18´ N) is located in interior of Illinois state, USA (Fig. 1b). A rectangular area (total area, 62,748 ha) is same with 33.2 km × 18.9 km. This area is considered part of the American corn belt; its major crops are soybean and maize (corn). Single-season cropping is generally practiced in this region. A detailed flowchart of the methods of this study is presented in Fig. 2.

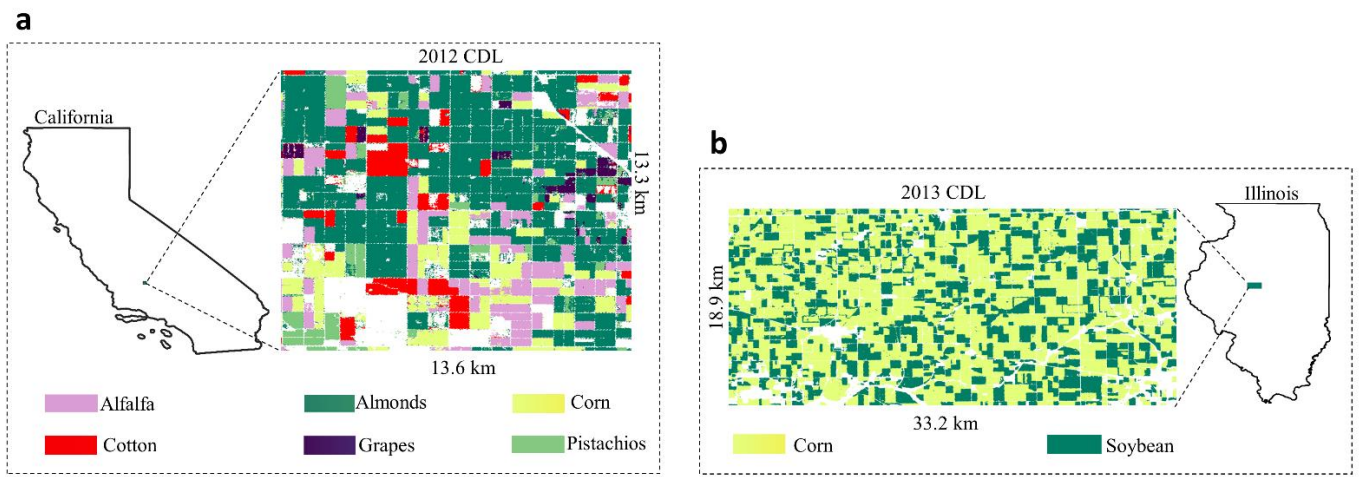

Fig. 1 Study area and major crop types in (a) California and (b) Illinois.

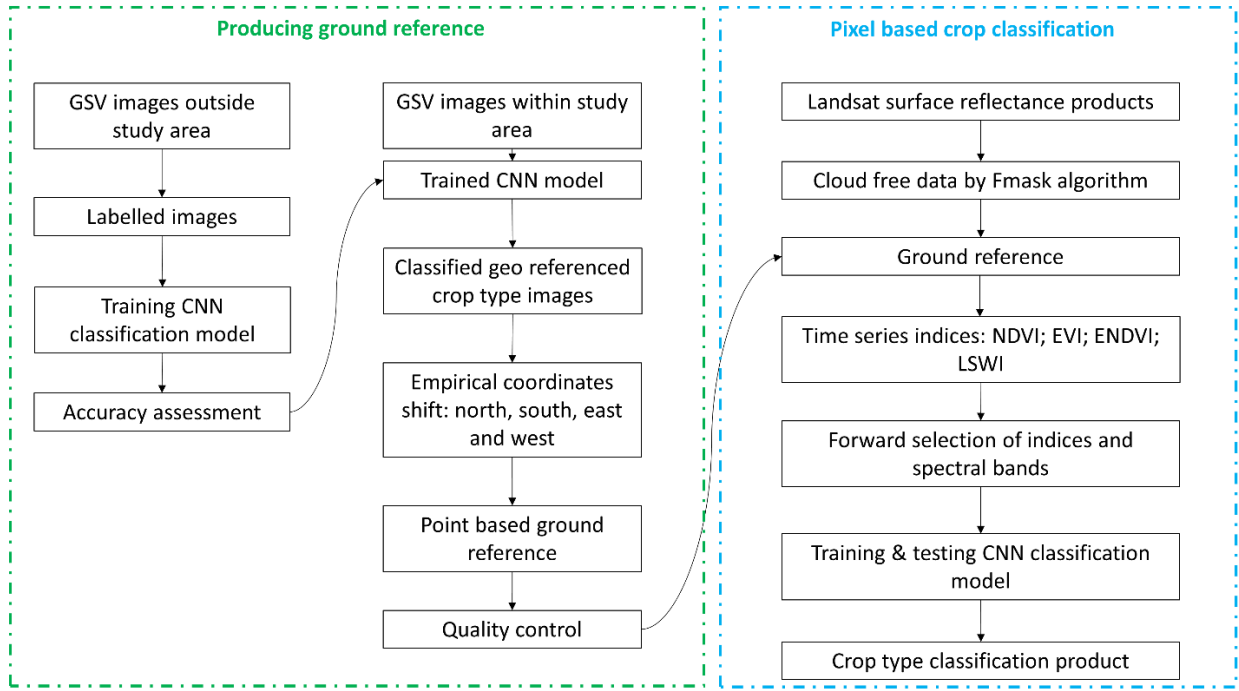

Fig. 2 Flowchart of the convolutional neural network (CNN)-based ground referencing and pixel-based crop type classification.

2.2 Google Street View (GSV) image collection

GSV images were automatically downloaded from Google Maps using an HTTP URL request using GSV API and a Python script. We searched for images by specifying latitude and longitude coordinates at intervals of around 30–60 m within the study areas. GSV provides panoramic images, to maximize information and minimize the number of images. We used a heading parameter to identify images facing four directions: north, east, south, and west. Time information was also retrieved for each image. Detailed information on the handling of GSV images is available at the Google Developer Guide (https://developers.google.com/maps/documentation/streetview/intro).

2.3 Training and validation of the CNN model

We applied a CNN model to classify all GSV images collected in the study areas into three classes in Illinois (corn, soybean and others), and seven classes in California (alfalfa, almond, corn, cotton, grape, pistachio and others). The "others" category included forest, grassland, man-made structures, and water bodies. Because most of the land cover in the study areas was identified as cropland, we did not further classify "others" according to land cover type. The key structure of the CNN model is presented in Fig. 3. To train the CNN model, we collected around 5,540 GSV images taken outside the study areas. We included images of different crop phenology stages, if available, especially for annual crops. The images were later randomly separated into three groups (60%, 20%, and 20%) for training (parameter fitting), validation (hyper-parameter tuning) and test (performance evaluation). To prepare training datasets for the CNN model, we manually classified GSV images into seven classes for California and three classes for Illinois (Fig. 4). To simplify this task, we targeted only ideal images containing a single homogeneous crop type (Fig. 4a–g). Images such as Fig 4i, containing both corn and forest, were classified as "others".

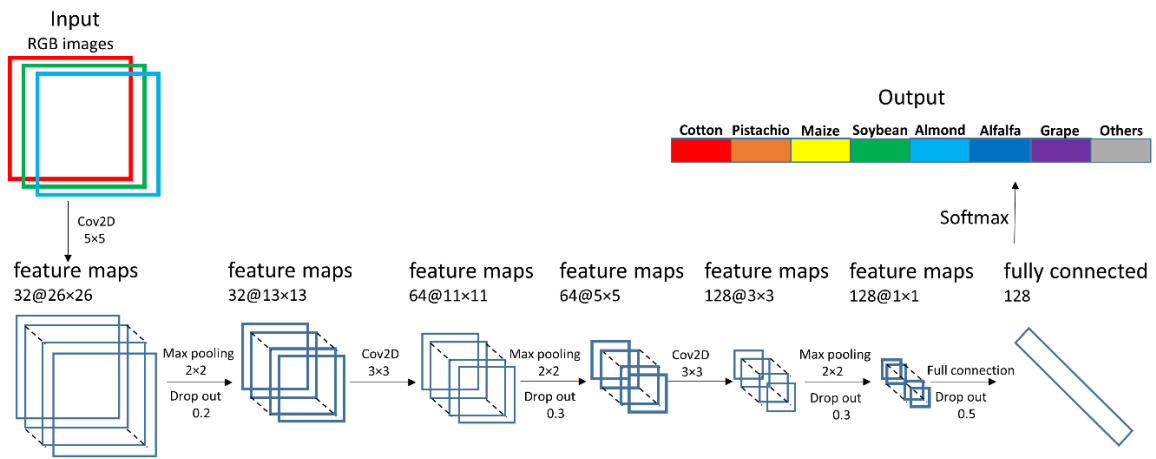

Fig. 3 Architecture of the CNN model for Google Street View (GSV) images classification. The feature map is the output of a single filter. Cov2D: two-dimensional convolution layer.

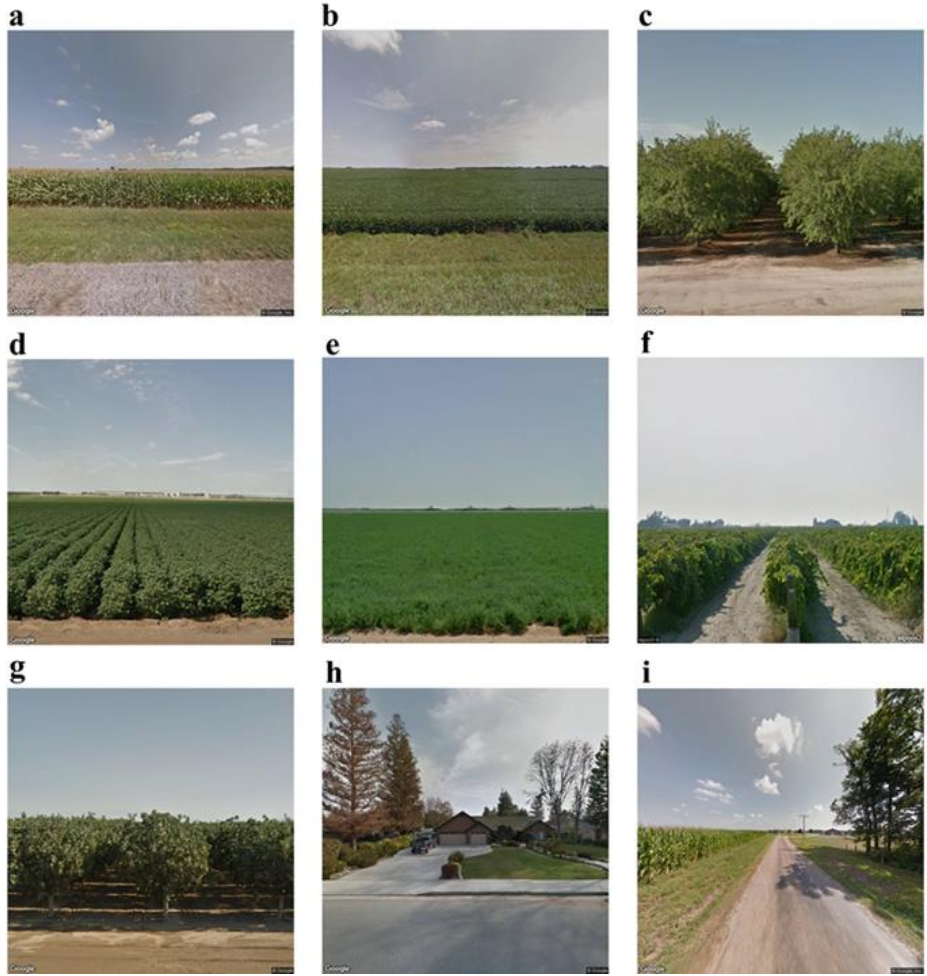

Fig. 4 Random GSV image samples used to train the CNN classification model; for (a) corn, (b) soybean, (c) almond, (d) cotton, (e) alfalfa, (f) grape, (g) pistachio, (h & i) others. Image copyright: Google Inc.

2.4 Producing ground reference data and quality control

We applied the CNN model to filter out crop images automatically from GSV images, which contained detailed geographic coordinate and time information. We first conducted quality control to remove minor images manually that were incorrectly classified by the CNN model. The coordinates of the image represent the location of the GSV vehicle, which is not identical to that of the crop parcel of interest (Fig. 5). Therefore, we shifted the coordinates from the GSV car to the captured parcel using the empirical coefficients x, y; an example of this process is shown in Fig. 5. We applied a buffer zone to avoid the reference points locating in the Landsat pixels mixing with road and crop parcels. Thus, the coordinates of targeted GSV images were moved 0.5y away from the car and around x away from a parcel edge (Fig. 5). If there were insufficient reference points for a certain crop type, we moved the coordinates several additional distance of 30 m in the same direction. The fixed empirical coefficient (x) value of 30 m between adjacent reference points was related to the spatial resolution of the Landsat 7 and 8 surface reflectance products used for mapping. For smaller parcel, a coefficient value of 10 m was more suitable to produce the reference points, if higher spatial resolution remote sensing products such as Sentinel were available. We used the same empirical coefficients for California and Illinois. To simplify the procedure, we considered only images captured from the four absolute cardinal directions: north, south, east and west. Ground reference points were validated using USDA CDL products and later used for crop types mapping.

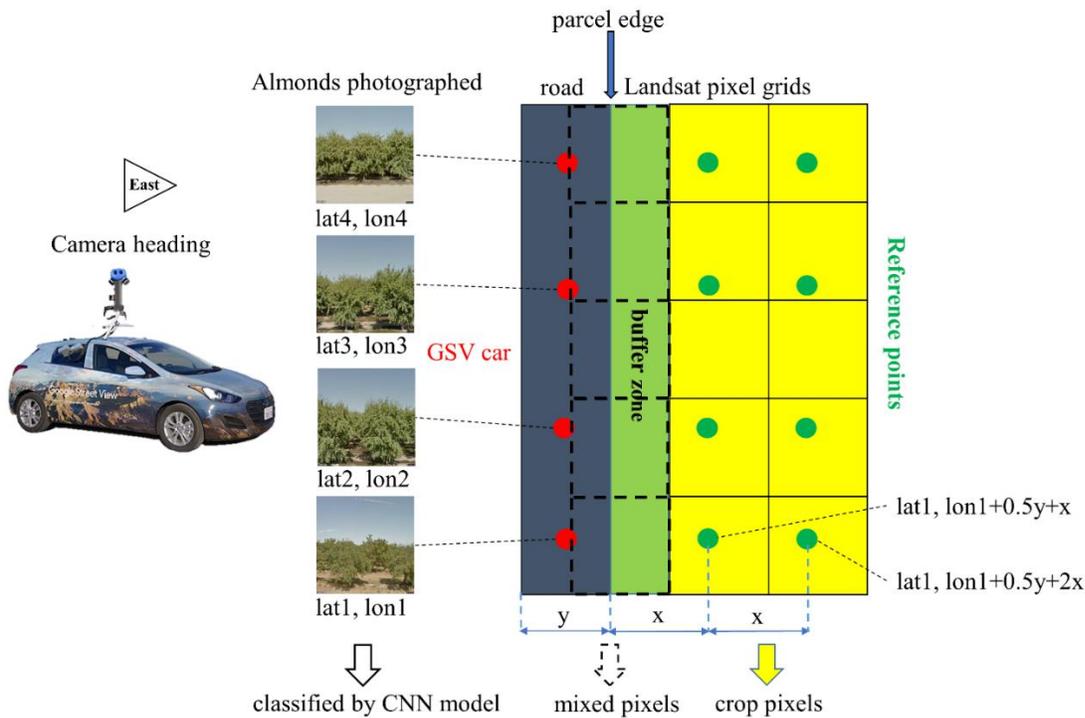

Fig. 5 GSV vehicle coordinates were empirically shifted to generate ground reference points. A buffer zone indicates potential mixed pixels near the parcel edge. Red dots indicate the position of GSV car; Green dots indicate reference points. y: road width; x: Landsat pixel resolution. Image copyright: Google Inc.

2.5 Mapping crop types

Images of remote crop parcels cannot be captured by GSV car, which are taken only from roads; therefore, crop mapping cannot rely solely on GSV images. It is also uncertain whether reference points generated near a parcel edge are robust for crop type mapping at the scale of the entire study area. Therefore, we performed pixel-based crop classification for Illinois and California using our reference points. We used the same CNN model (Fig. 3) and varied the dropout rate (0.1, 0.2, 0.3, and 0.5) and inputs to conduct crop classification. Temporal and spectral features from remote sensing surface reflectance products are commonly used to map crop types; however, different crop types with similar spectral information can hinder classification (Cai et al., 2018). We therefore combined time series spectral reflectance data (Red, Blue, Green, NIR, SWIR1 and SWIR2) and vegetation indices (VIs) (Eqs. 1–4) as input for the CNN model. We applied forward selection of the indices instead of testing all possibilities, because the total number of possible combinations was too large: $C_{10}^1 + C_{10}^2 + C_{10}^3 + \cdots\cdots + C_{10}^9 + C_{10}^{10} = 1023$. We maintained the variable that improved CNN model performance. The model was trained using 20 epochs each time. Forward selection was conducted for Illinois and we tested the generality of the final combination result by applying it to California, since more diverse crop types cultivated in California (Fig. 1a).

VIs were derived from Landsat 7 or 8 Collection 1 Level-2 scene products (surface reflectance) available at the United States Geological Survey database (https://earthexplorer.usgs.gov/). We applied the Fmask algorithm to remove data contaminated by clouds, cloud shadows and snow using the QGIS3 software (Zhu and Woodcock, 2012). We selected 19 scenes at path 24, row 32 and path 23, row 32 from March to October in 2013 for Illinois, and 14 scenes at path 42, row 35 from January to October in 2012 for California (Table 1). We computed the following vegetation indices: Normalized Difference Vegetation Index (NDVI); Enhanced Vegetation Index (EVI); Enhanced Normalized Difference Vegetation Index (ENDVI); Land Surface Water Index (LSWI), according to the following equations (Huete et al., 2002; LLC, 2014; Tucker, 1979; Xiao et al., 2002):

$$NDVI = (NIR - Red) / (NIR + Red) \qquad \text{Eq. 1}$$

$$EVI = 2.5 \times (NIR - Red) / (NIR + 6 \times Red - 7 \times Blue + 1) \qquad \text{Eq. 2}$$

$$ENDVI = ((NIR + Green) - (2 \times Blue)) / ((NIR + Green) + (2 \times Blue)) \qquad \text{Eq. 3}$$

$$LSWI = (NIR - SWIR1) / (NIR + SWIR) \qquad \text{Eq. 4}$$

To train the CNN model to map crop type, we randomly divided the reference points as 80% for training and 20% for validation. The mapping result was assessed using CDL products. The mapping accuracy of the CDL products ranged from 85–95% for major crop types (Boryan et al., 2011).

Table 1. Dates of Landsat 7 and 8 surface reflectance scenes used in this study.

| *Study area* | *Satellite* | *Year* | *Date* | | |
|---|---|---|---|---|---|
| *Illinois* | Landsat 8 | 2013 | Apr. (13, 20, 29) | May. (15, 22) | Jun. (7, 23) |
| | | | Jul. (9, 18, 25) | Aug. (3, 10, 19, 26) | Sep. (4, 11, 27) |
| | | | Oct. (6,13) | | |
| *California* | Landsat 7 | 2012 | Jan. (18) | Feb. (3) | Apr. (7, 23) |
| | | | May. (9, 25) | Jun. (10, 26) | Jul. (28) |
| | | | Aug. (13, 29) | Sep. (14, 30) | Oct. (16) |

## 3. Results and Discussion

3.1 GSV images classification

We assessed the accuracy of the CNN model using test datasets for both study areas (Fig. 6). The CNN model successfully classified GSV images in simple (three classes) and complex (seven classes) situations. In California, the overall accuracy reached 93%, and all producer accuracy results exceeded 90%, except for cotton (80%) and all user accuracy results exceeded 94%, except for corn (84%) and pistachios (86%). In Illinois, the CNN model performed better, with an overall accuracy of 97%, and producer and user accuracy results both exceeded 94%. The distinction of crop types in GSV images was generally robust but less effective at early phenological stages (e.g. in seedlings). We found that the seedling morphology was nearly identical across crop types due to low image resolution ($640 \times 640$). Neural network performance is susceptible to image quality (Dodge and Karam, 2016). Using higher resolution ($2048 \times 2048$) GSV images could further improve the model performance.

After accuracy assessment, we applied the CNN model to classify crops using GSV images. In California, we found 8,034 images belonging to seven classes: 374 alfalfa, 1,884 almond, 456 pistachio, 358 corn, 432 cotton, 88 grape and 4,442 others. In Illinois, we found 2,648 crop images for three classes: 1,044 corn, 720 soybean and 884 others. The CNN model occasionally confounded cotton, alfalfa, and corn (Fig.6a), especially for images taken during the early growing season. Therefore, we performed manual quality control to remove minor incorrectly classified crop images before producing the reference points. Another option is only using images taken around peak growing season during the model training process.

Fig. 6 Confusion matrices of classified GSV images at (a) California and (b) Illinois, PA: producer accuracy; UA: user accuracy; OA: the overall accuracy.

3.2 GSV images to ground reference

The ground reference was produced using classified GSV images via a coordinate shift to locate a certain crop type (Fig. 5). In total, we produced 8,173 crop reference points within parcels, 1,764 in Illinois and 6,409 in California (Fig. 7). It is also feasible to produce polygonal ground references by detecting the cropland parcels along the roads (e.g, edge extraction) (Graesser and Ramankutty, 2017) and connecting these with classified GSV images. This process is relevant for object-based mapping, especially in fragmented farming regions. To simplify the overall procedure, we produced reference points in this study.

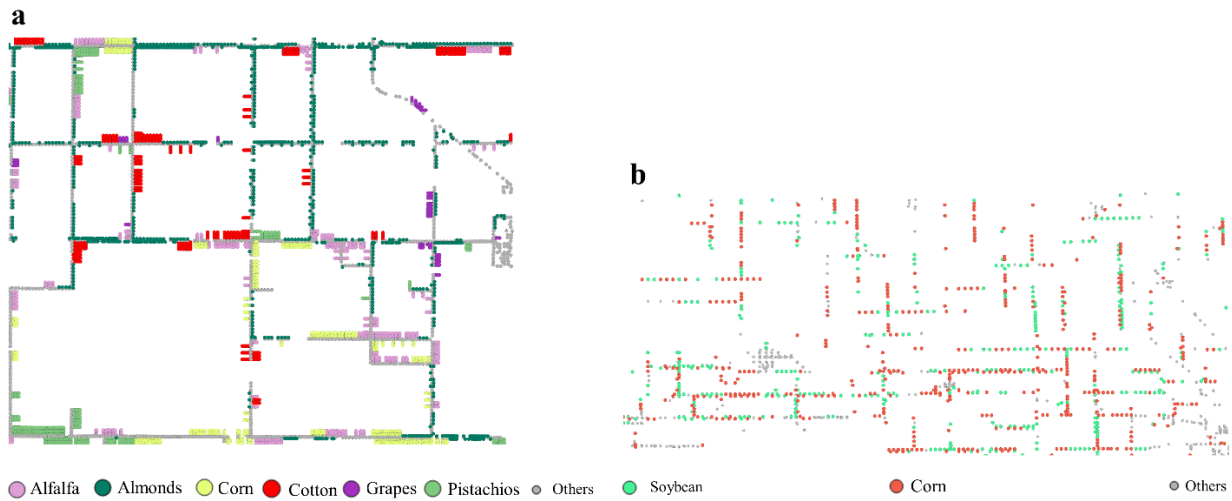

Fig. 7 Distribution of reference points derived from CNN and GSV images at (a) California and (b) Illinois.

3.3 Validation of the ground reference

We assessed the reliability of the GSV-derived crop type reference points using the USDA annual CDL products for California (2012) and Illinois (2013; Table 2). Ground reference points generally showed good agreement with the CDL products, at around 90% (1002/1115) for corn, 96% (1077/1120) for alfalfa, 89% (173/195) for grape, and 96% (955/994) for pistachio. In contrast, the results were more consistent for almond, 98% (1943/1984), and cotton 98% (980/1001). In Illinois, we found better agreement with CDL products compared to California, with only a few disagreements for corn (2%) and soybean (3%).

Table 2. Agreement between GSV derived ground reference points and CDL products.

| *California* | Alfalfa | Almond | Corn | Cotton | Grape | Pistachio |
|---|---|---|---|---|---|---|
| **Agreement** | 96% | 98% | 90% | 98% | 89% | 96% |
| *Illinois* | Corn | Soybean | | | | |
| **Agreement** | 98% | 97% | | | | |

3.4 Pros and cons of GSV images

We demonstrated the feasibility of producing ground reference points from GSV images using a deep learning model; we validated the model results using the CDL product. Conventional ground truthing methods, such as field surveys (Boryan et al., 2011), are time-consuming and costly. The CDL products are usually released 4–6 months after the growing season (Boryan et al., 2011; Cai et al., 2018). In contrast, the deep learning-GSV based approach is nearly full automatic and involves only minor non-automatic efforts during the post quality control phase. This approach provides an alternative for the timely producing ground references. We also collected around 20,000 GSV images for both CNN model training and reference production at a cost of 140 US $, which presents the potential for cost effective upscaling at the national scale.

GSV images provide considerable opportunities to produce ground reference points coherently at different regions. GSV, which was launched at 2007 in the USA, has been developed rapidly with international coverage in several major crop producing regions including North America, Europe, and parts of Asia and Latin America, with vast number of panorama images. However, image update frequency remains uncertain, on a scale of months to years, and also varies spatially. Images recorded during non-growing seasons usually provide little to no useful information unless special traits related to the specific crops are exhibited, such as standing rice stem residues after harvesting. GSV still has large data gaps in China, India, and Germany due to privacy concerns and local restrictions (Rakower, 2011). However, additional street view maps are available in these countries such as the Baidu map, Tencent map, and OpenStreetMap (Haklay and Weber, 2008; Liang et al., 2017; Long and Liu, 2017).

Recent trends in mining vast amounts of geo-tagged social media data for urban land use and tourist behavior studies could provide another potential solution for crop type ground truthing (Frias-Martinez and Frias-Martinez, 2014; Liu et al., 2017; Wood et al., 2013). Social media users continually generate recent land use information and share "big data" in the form of texts, images, or videos. These processes should be further investigated since these types of information are now far more timely and abundant than GSV images. A hybrid method that fuses GSV images, social media data, and census data, will be very valuable at the global level, for both crop type and land cover ground truthing.

3.5 Mapping using the ground reference

To demonstrate the reliability and usability of the GSV-derived ground reference, we conducted crop type mapping at the scale of the whole study areas. Due to the relatively small number of reference points, we randomly assigned 80% and 20% of the reference points for training and testing, respectively. The detailed validation accuracy for different combinations during the training phase is presented at Table 3. Adding the SWIR1 and SWIR2 spectra notably improved the mapping results. This finding is in agreement with a previous study in terms of corn and soybean classification (Cai et al., 2018) and demonstrated generality by improving a wide range of crop types classification, by including the SWIR1 and SWIR2 spectra. However, more model inputs did not necessarily lead to better performance. A model incorporates all VIs and spectral bands reached 90% accuracy, whereas higher accuracy (95%) was reached with fewer model inputs (EVI, ENDVI, SWIR1, SWIR2) (Table 3). Therefore, we chose the combination of EVI, ENDVI, SWIR1 and SWIR2 as the model inputs for mapping.

Table 3. Training accuracy of the CNN crop type classification model for the Illinois using different input combinations.

| MODEL INPUT | ACCURACY |
|---|---|
| EVI | 80% |
| ENDVI | 67% |
| NDVI | 75% |
| LSWI | 71% |
| EVI, ENDVI | 84% |
| EVI, NDVI | 82% |
| EVI, LSWI | 88% |
| EVI, SWIR1 | 83% |
| EVI, SWIR2 | 86% |
| EVI, ENDVI, SWIR1 | 85% |
| EVI, ENDVI, SWIR2 | 93% |
| EVI, SWIR1, SWIR2 | 94% |
| **EVI, ENDVI, SWIR1, SWIR2** | **95%** |
| EVI, NDVI, ENDVI, LSWI, RED, BLUE, GREEN, NIR, SWIR1, SWIR2 | 90% |

The mapping performance of the CNN model was evaluated using the ground reference points (Fig. 8). The CNN model was promising in its capacity for differentiating complex crop land cover types. In Illinois, the CNN model reached an overall accuracy of 94%. All user and producer's accuracies exceeded 90%. In California, the overall classification accuracy was 83%. Classification performance was lower, especially for grape (Fig. 8a). Due to the small cultivation area, limited grape reference points (88) were available to train the CNN model. Insufficient training samples dramatically reduce model performance (Foody et al., 2016; Kavzoglu, 2009). Despite advancements in the classifier, the quality and quantity of the reference for labelling crop types fundamentally affects classification accuracy, "garbage in, garbage out" (Foody et al., 2016; Kavzoglu, 2009). Minor land cover parcels were classified as "others", perhaps due to the treatment of mixed pixels as "others" during the training process.

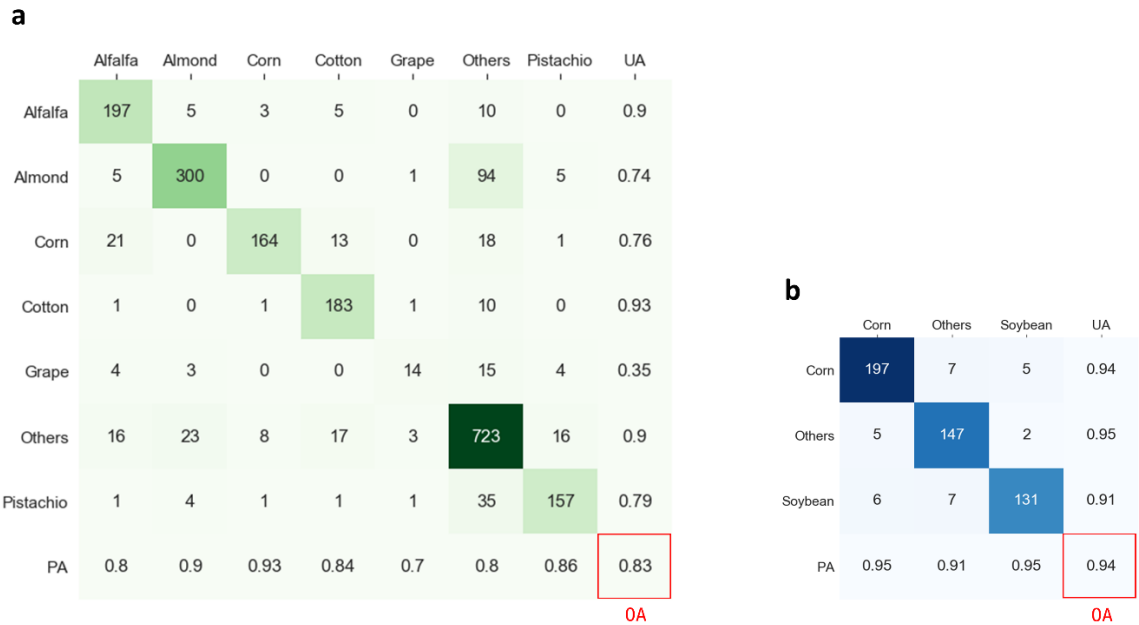

Fig. 8 Confusion matrices of crop type mapping at (a) California and (b) Illinois. PA: producer accuracy. UA: user accuracy. OA: the overall accuracy.

The mapping performance of CNN model was also assessed using the independent CDL products for 2012 (California) and 2013 (Illinois) at the scale of the entire study area (Figs. 9, 10, 11). The mapping accuracy of the CDL products ranged from 85–95% for major crop types (Boryan et al., 2011). The crop type mapping model captured the overall spatial pattern of land cover types. However, parcel edges were less distinct than those in the USDA CDL product, because the pixels with mixed corn and soybean were treated as "others" in this study. USDA CDL products are based on object-based classification, therefore, the mapping accuracy can be further improved by decomposing mixed pixels (Wang, 1990) and introducing image segmentation (Badrinarayanan et al., 2017); however, this work is beyond the scope of this study.

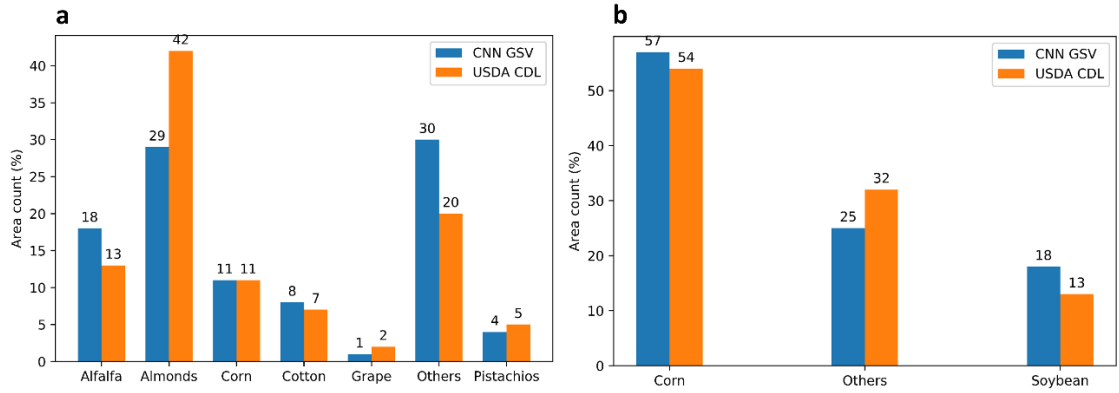

Fig. 9 Area counts of each land cover type in (a) California and (b) Illinois using USDA CDL and CNN based crop type classification.

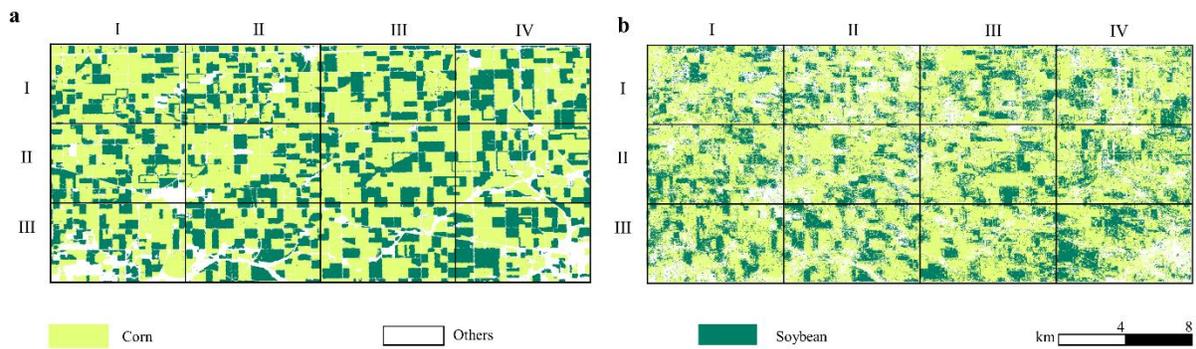

Fig. 10 Crop type mapping comparison for Illinois, using the (a) USDA cropland data layer product for 2013 and (b) CNN and GSV images derived map for 2013.

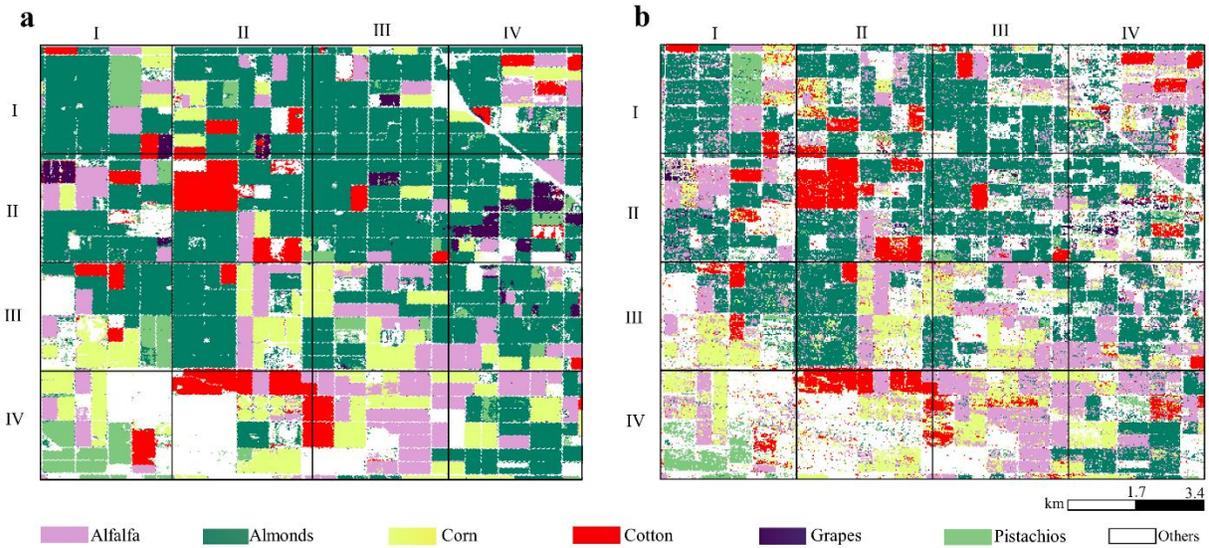

Fig. 11 Crop type mapping comparison for California, using the (a) USDA cropland data layer product for 2012 and (b) CNN and GSV images derived map for 2012.

## 4. Conclusion

Ground reference data are essential for supervised crop type mapping. This study provides a novel alternative method for nearly automated ground truthing by integrating a CNN model and GSV images. We demonstrated the general applicability of this new method by self-verification and evaluation with CDL products at two distinct farming areas. The CNN model showed considerable capability for GSV image classification and diverse crop type mapping. The ground reference points derived from GSV images taken along roads were representative and suitable for conducting regional-scale crop type classification. The high performance of the deep learning model presents upscaling potential for coherent ground truthing when GSV images are available.

## Acknowledgments

We thank Jungho Im for helpful comments on the manuscript. This work was conducted with the support of the Korea Environment Industry & Technology Institute (KEITI) through its Urban Ecological Health Promotion Technology Development Project, and funded by the Korea Ministry of Environment (MOE) (2019002760002). Proofreading service was offered by the Research Institute of Agriculture and Life sciences at SNU. Yulin Yan is also supported by the China Scholarship Council and SNU Global Scholarship.